\documentclass[10pt,twocolumn,letterpaper, table]{article}

\usepackage[pagenumbers]{cvpr} 
\usepackage{times}
\usepackage{latexsym}
\usepackage{algorithm}
\usepackage{algorithmic}
\usepackage{booktabs}
\usepackage{multirow}
\usepackage[T1]{fontenc}
\usepackage[utf8]{inputenc}
\usepackage{microtype}
\usepackage{amsmath}
\usepackage{amssymb}
\usepackage{float}
\usepackage{multicol}
\usepackage{booktabs}
\usepackage{xspace}
\usepackage{tcolorbox}
\usepackage{inconsolata}
\usepackage{booktabs}
\usepackage{pifont}
\usepackage{graphicx}
\usepackage[table, dvipsnames]{xcolor}

\newcommand{\ours}{Migician\xspace}
\newcommand{\ourtrain}{MGrounding-630k\xspace}
\newcommand{\ourbench}{MIG-Bench\xspace}
\newcommand{\cmark}{\ding{52}}
\newcommand{\hcmark}{\ding{52}\rotatebox[origin=c]{-9.2}{\kern-0.7em\ding{55}}}

\definecolor{mygreen}{RGB}{0,128,0}  
\definecolor{myred}{RGB}{255,0,0}

\definecolor{cvprblue}{rgb}{0.21,0.49,0.74}
\usepackage[pagebackref,breaklinks,colorlinks,allcolors=cvprblue]{hyperref}

\hypersetup{
    colorlinks=true
}
\def\paperID{*****} 
\def\confName{CVPR}
\def\confYear{2025}

\title{\raisebox{-0.27\height}{\includegraphics[width=0.35in]{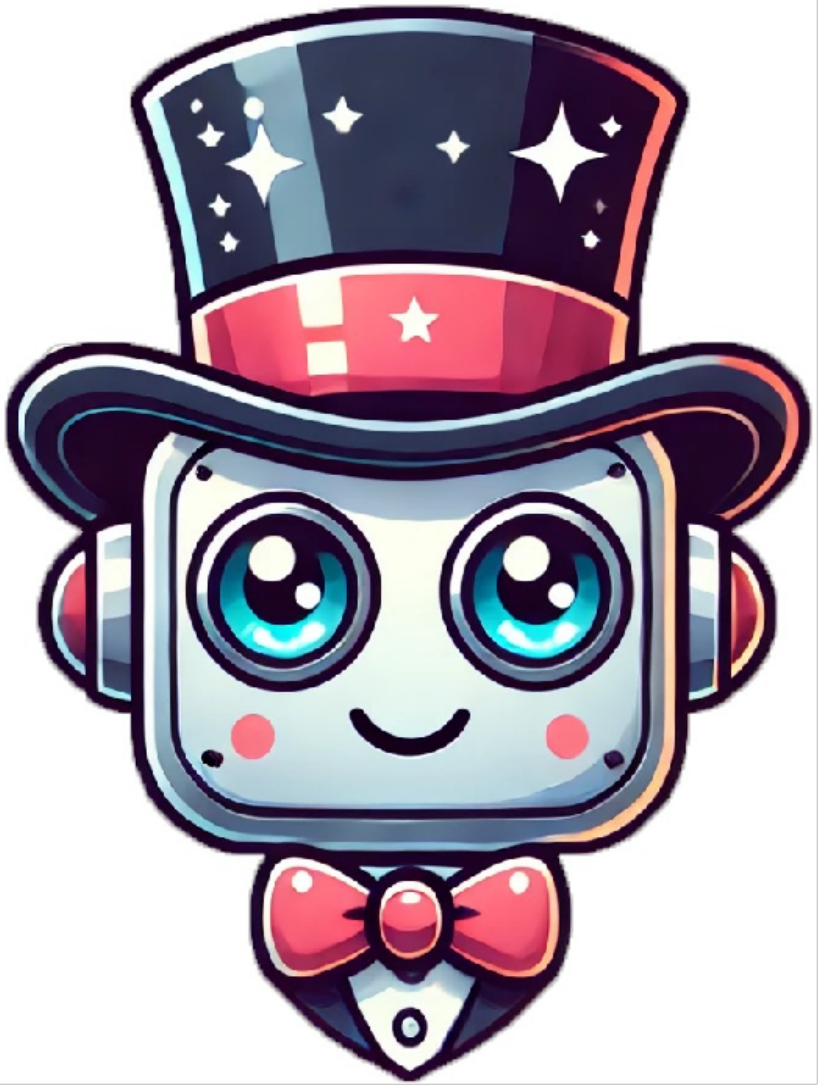}}~\ours: Revealing the Magic of Free-Form Multi-Image Grounding in Multimodal Large Language Models}

\author{
    \textbf{You Li\textsuperscript{1}\footnotemark[1]},
    \textbf{Heyu Huang\textsuperscript{2}\footnotemark[1]},
    \textbf{Chi Chen\textsuperscript{3}\footnotemark[2]},
    \textbf{Kaiyu Huang\textsuperscript{1}\footnotemark[2]}, 
    \textbf{Chao Huang\textsuperscript{1}},
    \textbf{Zonghao Guo\textsuperscript{3}},
    \\
    \textbf{Zhiyuan Liu\textsuperscript{3}}, 
    \textbf{Jinan Xu\textsuperscript{1}},
    \textbf{Yuhua Li\textsuperscript{2}},
    \textbf{Ruixuan Li\textsuperscript{2}},
    \textbf{Maosong Sun\textsuperscript{3}}
    \\
\textbf{\textsuperscript{1}} State Key Laboratory of Advanced Rail Autonomous Operation, \\
Beijing Jiaotong University, Beijing, China
\\
\textbf{\textsuperscript{2}} Huazhong University of Science and Technology, Wuhan, China
\\ 
\textbf{\textsuperscript{3}} Tsinghua University, Beijing, China 
}

\begin{document}

\twocolumn[{
\maketitle
\vspace{-25pt}
\begin{center}
    \centering
    \captionsetup{type=figure}
    \includegraphics[width=\textwidth]{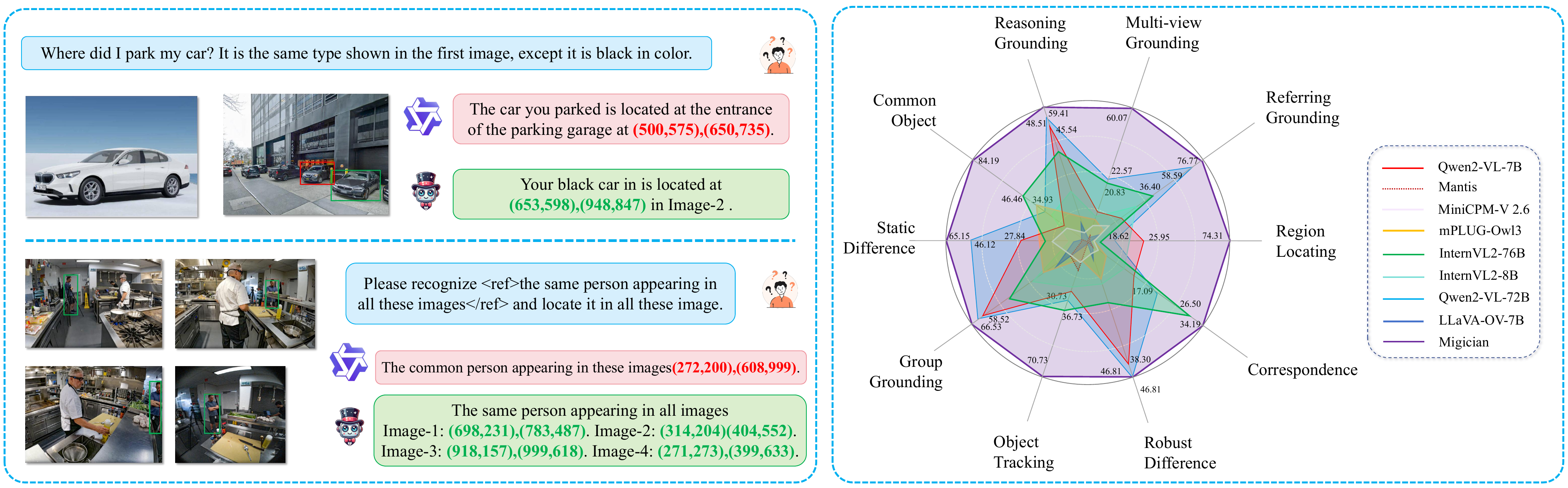}
  \vspace{-2.0em}
  \caption{\textbf{Left}: Examples of free-form multi-image grounding. The task is to identify and localize relevant visual regions across multiple images based on a free-form query. \textbf{Right}: Our proposed model, \ours, significantly outperforms other MLLMs on various multi-image grounding tasks.}
  \label{fig:radar}
\end{center}
}]

\renewcommand{\thefootnote}{\fnsymbol{footnote}}
\footnotetext[1]{Equal contribution.}
\footnotetext[2]{Corresponding authors: Chi Chen (chenchithu@gmail.com) and Kaiyu Huang (kyhuang@bjtu.edu.cn).}

\begin{abstract}
The recent advancement of Multimodal Large Language Models~(MLLMs) has significantly improved their fine-grained perception of single images and general comprehension across multiple images. However, existing MLLMs still face challenges in achieving precise grounding in complex multi-image scenarios.
To address this, we first explore a Chain-of-Thought~(CoT) framework that integrates single-image grounding with multi-image comprehension. While partially effective, it remains unstable and struggles to capture abstract visual information due to its non-end-to-end nature. Therefore, we introduce \textbf{\ours}, the first multi-image grounding model capable of performing free-form and accurate grounding across multiple images. To support this, we present the \ourtrain dataset, which comprises data for several multi-image grounding tasks derived from existing datasets, along with newly generated free-form grounding instruction-following data. Furthermore, we propose \ourbench, a comprehensive benchmark specifically designed for evaluating multi-image grounding capabilities.
Experimental results demonstrate that our model achieves significantly superior multi-image grounding capabilities, outperforming the best existing MLLMs by 24.94\% and even surpassing much larger 70B models. Our code, model, dataset, and benchmark are fully open-sourced at \textit{\href{https://migician-vg.github.io/}{https://migician-vg.github.io/}}.
\end{abstract}

\section{Introduction}
\label{sec:intro}

Multimodal Large Language Models~(MLLMs) have exhibited significant advancements recently, demonstrating exceptional cross-modal understanding capabilities and achieving outstanding performance in various vision-language tasks~\cite{ye2023ureader, hu2024mplug, elliott2017imagination, ive2019distilling, lu2021iconqa, amini2019mathqa, krishna2017visual}. As these models continue to evolve, their capabilities have expanded beyond image-level understanding to include fine-grained visual grounding~\cite{wang2023one,chen2023shikra,you2023ferret}. 
This enables MLLMs to process region-specific inputs and outputs, unlocking a broader spectrum of real-world multimodal application scenarios~\cite{peng2023kosmos}.

Despite the promising visual grounding capabilities demonstrated by existing MLLMs, these abilities are largely confined to single-image scenarios~\cite{kazemzadeh2014referitgame, you2023ferret}. The potential of MLLMs in free-form \textbf{multi-image grounding~(MIG)} remains underexplored. Free-form MIG challenges the model to perform grounding across multiple images effectively, where the input queries and image contexts can be organized in arbitrary forms, enabling flexible and dynamic interactions. For instance, as shown in Figure~\ref{fig:radar}, the model must understand the white car in the query image and relate it to the textual prompt "black in color" to identify the corresponding target in the target image.
This capability unlocks a wide range of applications, such as fine-grained environmental perception in autonomous driving~\cite{wang2024driving}, anomaly detection in surveillance systems~\cite{black2002surveillance}, and target localization for embodied robotics~\cite{grauman2022ego4d}.
To address the free-form MIG, the model needs to possess the capability for visual grounding while achieving cross-image understanding.

As a result, a question naturally arises: \textit{Can we integrate the single-image grounding and multi-image understanding capabilities of existing MLLMs to tackle the MIG task?} In this work, we propose a Chain-of-Thought~(CoT) framework that first leverages multi-image understanding to generate a textual referring query, and then utilizes it for localization through single-image grounding. This approach has been proven highly effective for MIG tasks, particularly in simple scenarios where textual descriptions are sufficiently distinctive, demonstrating the potential of current MLLMs in handling such tasks. 

However, the proposed CoT framework struggles with describing abstract visual semantics in multi-image scenarios, and the two-step process results in a doubling of the inference time. To address this, we further propose Migician, a competitive MLLM capable of free-form and accurate grounding across multiple images, serving as an end-to-end solution for MIG.
To progressively establish flexible grounding capabilities, we employ a two-stage training procedure based on our proposed large-scale MIG dataset~(\ourtrain). First, the grounding ability of Migician is enhanced through a combination of data from MIG tasks and general tasks. Then, Migician is further refined using high-quality free-form MIG instruction data.
In addition, to evaluate the challenges of the free-form MIG scenario, we construct a comprehensive multi-image grounding benchmark, MIG-bench, comprising a total of 10 different tasks, 5.9k diverse images and more than 4.2k test instances.
We observe a significant gap between the performance of existing mainstream MLLMs and human performance on the MIG-bench. In contrast, Migician can effectively alleviate this gap and improve the performance of free-form MIG.

To sum up, our contributions can be concluded as follows:
\begin{itemize}
\item We explore the task of multi-image grounding for MLLMs and  reveal the potential and challenges of current MLLMs by through a proposed CoT framework.

\item We introduce \ours, the first MLLM capable of effectively performing free-form MIG. We also present \ourtrain, the first large-scale MIG instruction tuning dataset for training this model. 

\item We introduce \ourbench, a comprehensive benchmark for evaluating multi-image grounding capabilities. Experimental results demonstrate that \ours significantly outperforms the current best methods.
\end{itemize}

\section{Related Work}
\label{sec:Related}

\paragraph{Multimodal Large Language Models} 
Recent developments in multimodal large language models (MLLMs) have shifted from single image-text understanding towards more versatile capabilities~\cite{cai2024internlm2,yao2024minicpm,wang2024qwen2,li2024llavaov}. Among these efforts, some focus on enabling models to achieve fine-grained visual grounding, either through simple instruction tuning~\cite{chen2023shikra,peng2023kosmos} or by integrating additional auxiliary visual components~\cite{you2023ferret,zhang2023gpt4roi,chen2023position}. However, these models primarily focus on visual grounding within a single image. Some other studies explore multi-image understanding tasks, such as multi-image comparison, reasoning, and temporal comprehension~\cite{jiang2024mantis, li2024llavainterleave,ye2024mplug,li2024llavaov,cai2024internlm2,yao2024minicpm}. Nevertheless, fine-grained visual grounding at the multi-image level remains an underexplored area. To the best of our knowledge, our proposed \ours is the first MLLM designed to address the challenge of multi-image grounding.

\paragraph{MLLM Benchmarks} 
Most existing benchmarks for evaluating MLLMs focus on single-image tasks~\cite{fu2023mme,li2024seed}. A few recent benchmarks have started assessing the performance of MLLMs on multi-image understanding~\cite{jiang2024mantis,meng2024mmiu,fu2025blink,wang2024muirbench,liu2024mibench}, but they primarily emphasize image-level comprehension.
The most relevant benchmark to our work is MC-Bench~\cite{xu2024mc}, a contemporaneous study. MC-Bench evaluates the multi-context grounding capabilities of MLLMs by asking them to accurately locate the corresponding object based on a text prompt in the correct image from a given pair. However, it exhibits limitations in the fixed number of input images and the restricted forms of queries. In contrast, the proposed \ourbench in this work offers more flexible task formats, focusing on evaluating models' capabilities in free-form multi-image understanding.

\begin{figure*}[h] 
    \centering
    \includegraphics[width=\textwidth]{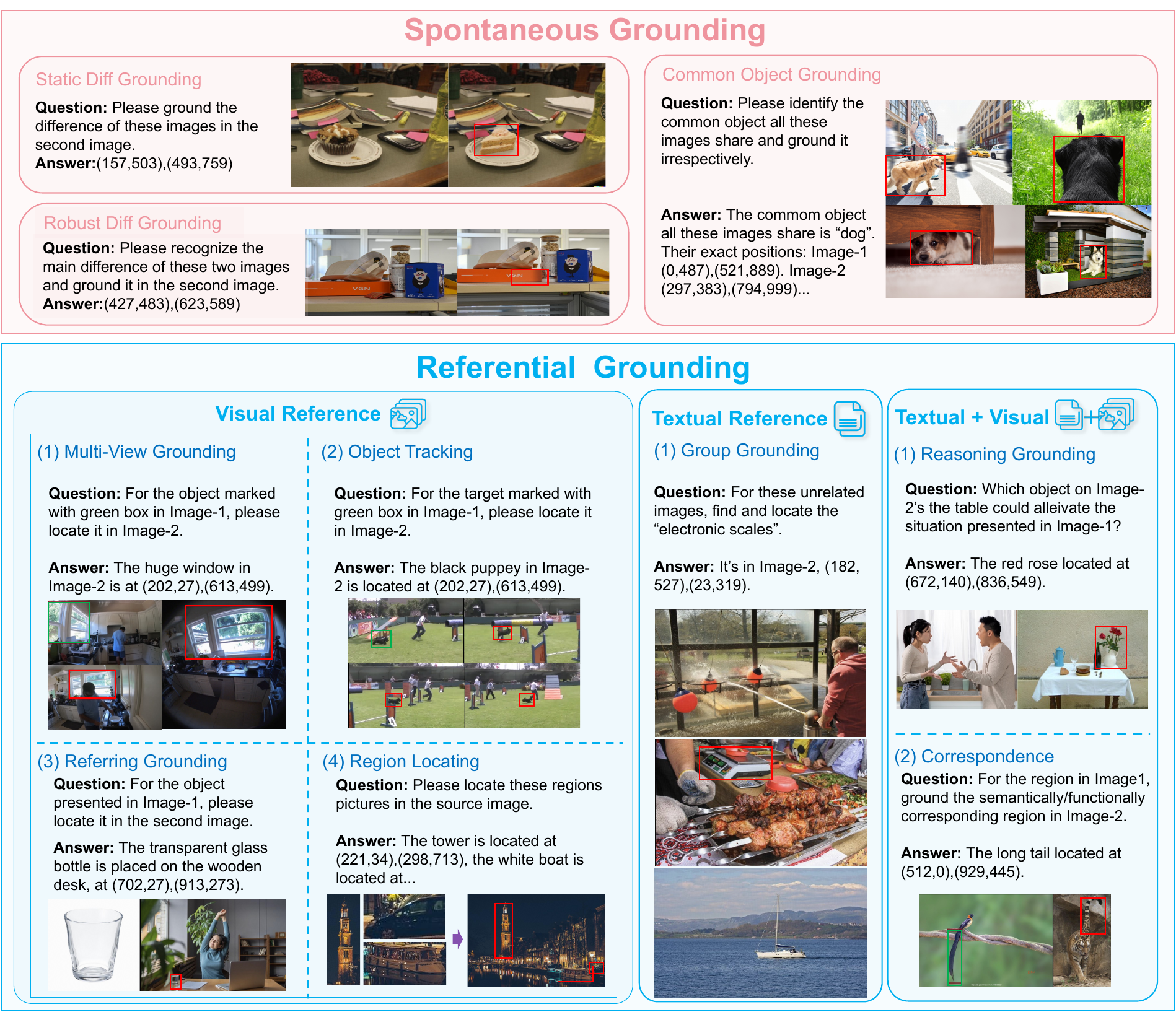} 
    \caption{
An illustration of the multi-image grounding tasks included in \ourbench. These tasks are divided into two categories: spontaneous grounding and referential grounding, depending on the whether there are explicit referential requirements.} 
    \label{fig:MIGtasks} 
\end{figure*}

\section{Task Definition}
The task of free-form multi-image grounding is to identify and localize relevant visual regions across a set of images based on a free-form query. Unlike traditional grounding tasks with fixed input formats, the query in free-form multi-image grounding can be an \textbf{arbitrary combination of text and images}, making it highly flexible and versatile. Formally, let the query $Q$ consist of a natural language description, reference images $\{R_1, R_2, \ldots, R_k\}$ or a hybrid combination of both (e.g., ``[a white car image] find a car like this image except it is black''). Given a set of target images $\{I_1, I_2, \ldots, I_n\}$, the task is to identify a set of visual regions $\{G_1, G_2, \ldots, G_m\}$ where $G_i$ is the target region within an image $I_j$ that satisfies the semantic and contextual constraints defined by $Q$. 
\begin{figure*}[t] 
    \centering
    \includegraphics[width=\textwidth]{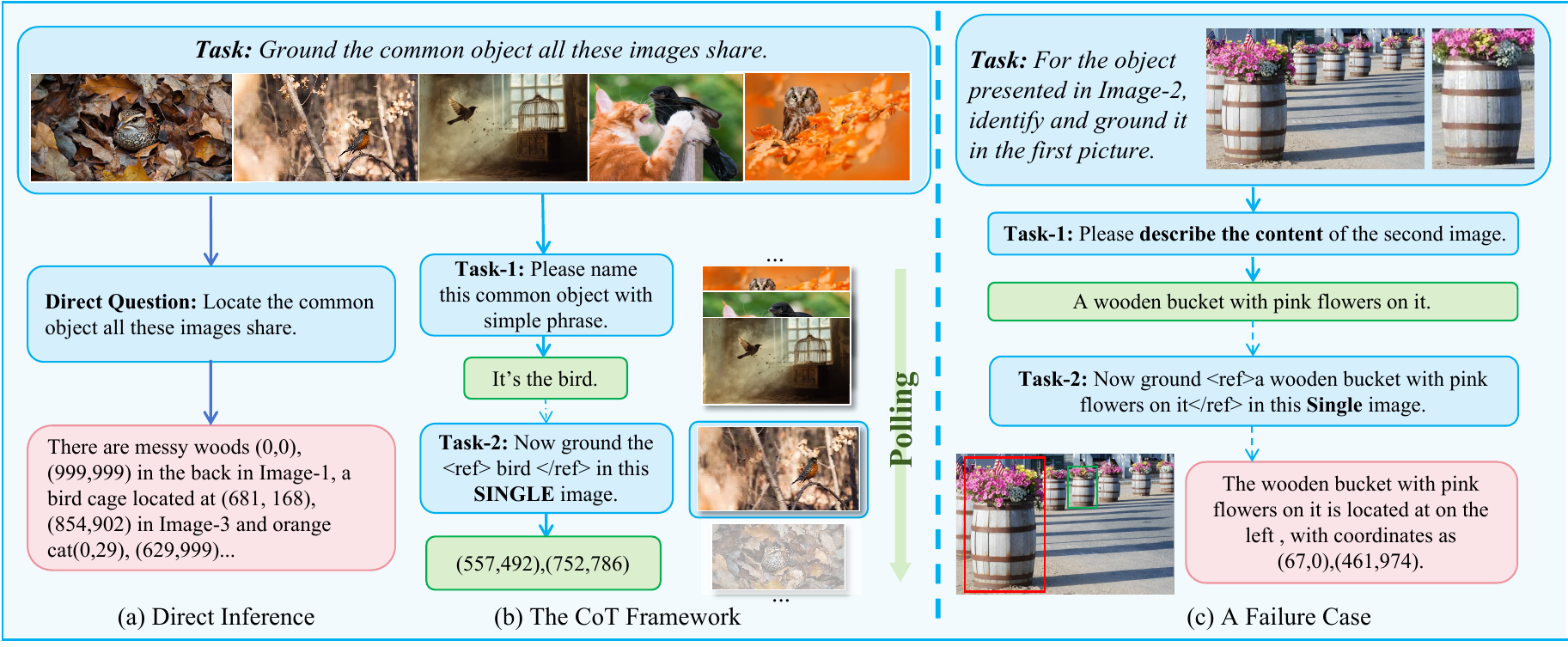} 
    \caption{Illustration of the CoT framework and its failure case. Different from (a) direct inference, the (b) CoT method decomposes the task into two subtasks, solving each task deploying the model’s existing capabilities. A failure case of CoT is shown in (c) where the model struggles at handling abstract visual information. Green and red background colors indicate correct and incorrect answers, respectively.} 
    \label{fig:cot_illustration} 
\end{figure*}
As shown in Figure~\ref{fig:MIGtasks}, based on whether the task involves explicit reference requirements, multi-image grounding tasks can be further categorized into two types: \textit{Spontaneous Grounding} and \textit{Referential Grounding}. Spontaneous Grounding refers to recognizing and grounding the target object in corresponding images without explicitly pointing it out.
Unlike the conventional Reference Expression Comprehension task~\cite{kazemzadeh2014referitgame} that explicitly refer to the target object, Spontaneous Grounding typically utilizes the relationships between multiple images as contextual cues to autonomously identify and localize the objects to be grounded~(e.g., finding and locating differences between images). Referential Grounding, on the other hand, requires an explicit reference to the target object. As mentioned earlier, such references can take the form of arbitrary combinations of images and textual descriptions.

\section{Methods}


\begin{figure*}[t] 
    \centering
    \includegraphics[width=\textwidth]{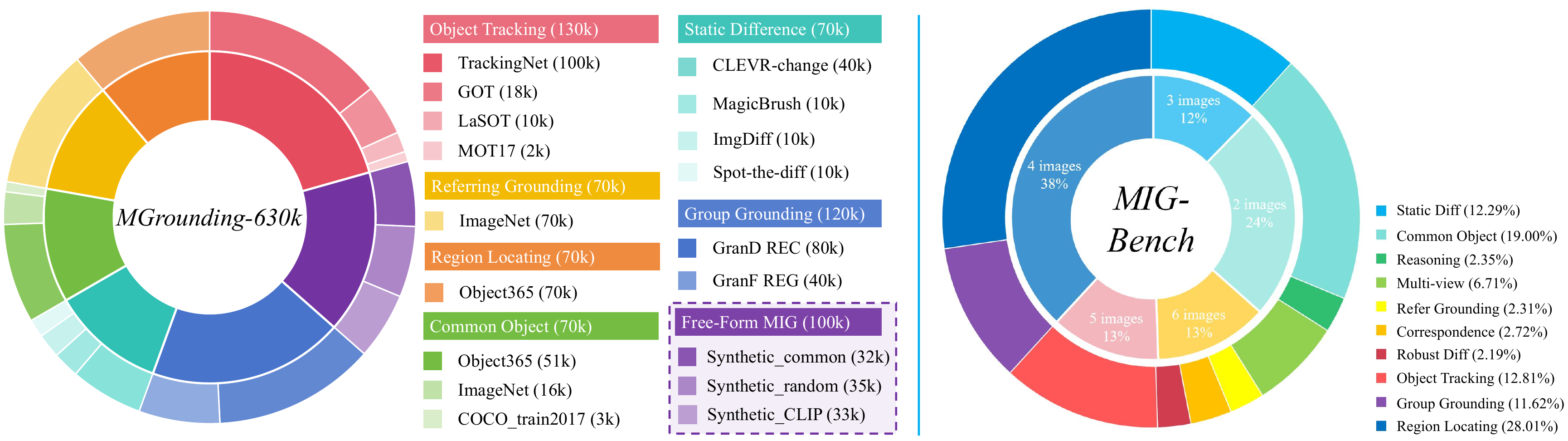} 
    \caption{Statistics of the \ourtrain dataset and \ourbench.} 
    \label{fig:statistics} 
\end{figure*}

In this section, we delve into methods for enabling free-form multi-image grounding capabilities in MLLMs. We begin by exploring a Chain-of-Thought~(CoT) framework to elicit the capabilities within existing MLLMs to tackle this task. Then  we develop an end-to-end MIG model, \ours, through instruction tuning to overcome the limitations of the CoT framework and achieve enhanced MIG performance.

\subsection{A Chain-of-Thought Framework}
\label{sec:cot}
As illustrated in Figure~\ref{fig:cot_illustration}(a), directly prompting existing MLLMs to perform MIG tasks often leads to significant performance degradation. To better explore the potential of existing models, 
we design a CoT framework which decouples the MIG task into two stages as shown in Figure~\ref{fig:cot_illustration}(b). The model first understands the input images and question, generating a textual referring expression that describes the target object. Next, the model locates the objects in corresponding images using the referring expression from the previous step.

This framework leads to a notable performance improvement on MIG tasks. However, the CoT framework has inherent limitations, such as error propagation due to its multi-step process, which also reduces reasoning efficiency~\cite{yao2022react}. Additionally, as illustrated in Figure~\ref{fig:failure}(c), many scenarios require grounding through abstract visual semantics across multiple images, which cannot be effectively captured through textual expressions. More failure types are detailed in Appendix~\ref{appendix:failure}. This highlights the need for an end-to-end model capable of directly performing the MIG task.

\subsection{Data Construction}

The CoT framework has demonstrated that an MLLM with both multi-image understanding and single-image grounding capabilities inherently holds strong potential for free-form MIG. In the following section, we employ instruction tuning to explicitly bridge these capabilities in existing MLLMs to achieve MIG. For this purpose, we first construct an instruction tuning dataset for MIG, named \textbf{\ourtrain}, with its statistics presented in Figure~\ref{fig:statistics}. This dataset is primarily constructed through the following two ways.

\paragraph{Transforming Existing Data.} By analyzing the tasks and annotation types of existing datasets, we identify multiple multi-image grounding (MIG) tasks whose data could be derived through transformation of the existing. Specifically, we collect and organize data from existing sources, combining or automatically synthesizing single-image annotations to create datasets for 6 types of MIG tasks. Each task contains over 70k examples, resulting in a total of 530k training samples. The details of these task data are in Appendix \ref{appendix:stage1data}.

\paragraph{Synthesizing Free-form MIG Data.} 
The data obtained through the aforementioned methods still do not fully meet the requirements for free-form MIG. To acquire MIG data with richer and more diverse formats, which would enhancing the model's instruction-following and flexible grounding capabilities, we design a MIG data synthesis pipeline. This pipeline uses the Object365~\cite{shao2019objects365} images with object annotations, select multiple images as a group, and generate high-quality instructions for multi-image grounding. 
Specifically, we first employ Qwen2-VL-72B~\cite{wang2024qwen2} to generate captions of each individual image and then perform error filtering and refinement on the annotated bounding boxes. Next, we prompt Qwen2.5-72B~\cite{yang2024qwen2} to automatically generate high-quality, free-form MIG question-answering pairs by integrating information from multiple images. 
To optimize the selection of appropriate image groups, we adopt different image grouping methods, including random selection, selection of images with common objects, and grouping images based on CLIP similarity to select semantically similar images for each. Using these methods, we generate a total of 100k Free-Form MIG data. Detailed information can be found in Appendix~\ref{appendix:stage2data}.

\subsection{Instruction Tuning for MIG}

Using the constructed dataset, we perform instruction tuning based on Qwen2-VL-7B~\cite{wang2024qwen2} to develop \textbf{\ours}, enabling it to achieve end-to-end free-form MIG capabilities.

\paragraph{Two-Stage Training.} To effectively equip the model with free-form MIG capabilities, we propose a two-stage training approach. In the first stage, the model learns to perform multi-image grounding by training on the six representative MIG tasks of \ourtrain, acquiring the ability to simultaneously comprehend multiple images and execute visual grounding. In the second stage, the model is further fine-tuned on free-form MIG instruction data in \ourtrain, enabling it to adapt to more flexible and diverse instruction types and transfer the MIG skills learned in the first stage to a broader range of scenarios. To prevent the model from forgetting its existing capabilities during training, we also incorporate single-image understanding, multi-image understanding, and single-image grounding data into each training stage. For more details please refer to the Appendix~\ref{appendix:twostage}.

\begin{table*}[t] 
\centering

\resizebox{\textwidth}{!}{
    \begin{tabular}{l|ccc | ccccccc|c}
    \toprule
    \multirow{4}{*}{\textbf{Models}} & \multicolumn{3}{|c}{\textbf{Spontaneous Grounding }} & \multicolumn{7}{|c|}{\textbf{ Referential Grounding }} & \multirow{4}{*}{\textbf{AVE}} \\ 
    \cmidrule(lr){2-4} \cmidrule(lr){5-11} 
    
    \multicolumn{1}{c}{} & \multicolumn{2}{|c|}{\textbf{Difference}} & \textbf{Similarity} &  \multicolumn{4}{c|}{\textbf{Visual Reference}} & \textbf{Textual} & \multicolumn{2}{|c|}{\textbf{Visual+Textual}}& \\ 
    \cmidrule(lr){2-3} \cmidrule(lr){4-4} \cmidrule(lr){5-8} \cmidrule(lr){9-9} \cmidrule(lr){10-11}
    
    \multicolumn{1}{c|}{} & \textbf{Static} & \textbf{Robust} & \multicolumn{1}{|c|}{\textbf{Common}}& \textbf{OT} & \textbf{MV} & \textbf{Region} & \textbf{Refer} &  \multicolumn{1}{|c|}{\textbf{GG}} & \textbf{Reason} & \multicolumn{1}{c|}{\textbf{Co-Re}} & \\

    \midrule
    
    \rowcolor[HTML]{B0E0E6}
    \multicolumn{12}{c}{\textbf{Human Performance}}\\
    \midrule Human & 99.50* & 97.87 & 98.00* & 100.00 & 96.88 & 100.00* & 98.99 & 91.06* & 92.08 & 97.44 & 97.18
    \\
    \midrule 

    \rowcolor[HTML]{ACE8A1}
    \multicolumn{12}{c}{\textbf{70B-Scale MLLMs}}\\
    \midrule

    LLaVA-OV-72B & 13.26 & 5.34 & 26.84 & 12.91 & 7.64 & 2.14 & 17.83 & 21.60 & 11.88 & 8.55 & 13.65 \\
    InternVL2-76B & 15.91 & 10.64 & 36.40 & 30.73 & 20.83 & 5.74 & 46.46 & 41.28 & 32.67 & 26.50 & 26.72 \\
    Qwen2-VL-72B & 46.12 & 46.81 & 64.46 & 26.73 & 22.57 & 18.62 & 33.33 & 62.53 & 50.50 & 17.09 & 38.88 \\
    
    \midrule 
    \rowcolor[HTML]{F0D5A2}
    \multicolumn{12}{c}{\textbf{7B-Scale MLLMs}}\\
    \midrule
    Mantis & 1.52 & 0.00 & 3.31 & 12.18 & 2.08 & 1.00 & 1.01 & 10.02 & 0.00 & 0.85 & 3.20 \\
    LLaVA-OV-7B & 6.06 & 3.19 & 3.43 & 0.18 & 1.04 & 1.08 & 9.09 & 15.43 & 6.93 & 0.85 & 4.73 \\
    Minicpm2.6 & 14.58 & 2.13 & 14.34 & 9.82 & 6.25 & 1.75 & 11.11 & 10.02 & 2.97 & 2.56 & 7.55 \\
    mPLUG-Owl3 & 18.56 & 6.38 & 34.93 & 8.55 & 7.64 & 2.41 & 7.07 & 22.85 & 9.09 & 5.98 & 12.35 \\
    InternVL2-8B & 6.92 & 7.45 & 25.49 & 20.73 & 9.72 & 3.49 & 28.28 & 30.26 & 17.82 & 9.40 & 15.96 \\
    Qwen2-VL-7B & 27.84 & 38.30 & 19.36 & 20.73 & 11.81 & 25.95 & 23.23 & 58.52 & 48.51 & 11.97 & 28.62 \\
    mPLUG-Owl3$_{+CoT}$ & 16.29 & 8.51 & 55.39 & 44.36 & 25.35 & 19.04 & 36.36 & 30.86 & 18.81 & 10.26 & 26.52 \\
    InternVL2-8B$_{+CoT}$ & 14.58 & 7.45 & 72.54 & 40.91 & 27.78 & 28.60 & 67.68 & 44.49 & 41.58 & 11.97 & 35.76 \\
    Qwen2-VL-7B$_{+CoT}$ & 23.48 & 40.43 & 63.85 & 62.73 & 42.71 & 24.85 & 54.55 & 43.29 & 51.49 & 30.77 & 43.82 \\
    
    \midrule
    Migician & \textbf{65.15} & \textbf{46.81} & \textbf{84.19} & \textbf{70.73} & \textbf{60.07} & \textbf{74.31} & \textbf{76.77} & \textbf{66.53} & \textbf{59.41} & \textbf{34.19} & \textbf{63.82} \\
    
    \bottomrule
    \end{tabular}
}
\caption{Performance comparison of different models on MIG-Bench. OT, MV, GG and Co-Re respectively means object tracking, multi-view grounding, group grounding and correspondence. For values marked with *, we randomly sample 20\% testing examples for human evaluation on the corresponding task.}
\label{tab:mig-bench}
\end{table*}

\paragraph{Model Merging.} After the second stage of fine-tuning, we observe a trade-off between model performance and flexibility: while the model adapts to the free-form MIG instructions, there is a performance drop in common multi-image grounding tasks. To better balance these two aspects, we adopt the model merging technique~\cite{ilharco2022editing}, averaging the model weights obtained from both training stages as the final weights. We find this approach mitigates the performance loss in common MIG tasks while preserving the ability to follow free-form MIG instructions effectively.

\section{MIG-Bench}
We introduce \ourbench, a manually curated benchmark designed to evaluate the MIG ability of current MLLMs. It comprises 5.9k images, and 4.3k testing instances, covering 10 distinct tasks shown in Figure~\ref{fig:MIGtasks} with details in Appendix~\ref{appendix:Benchmark Definition}. The distribution of these tasks is illustrated in Figure~\ref{fig:statistics}. 

\ourbench is manually constructed from multiple data sources. Initially, we select annotated data examples from existing datasets and adapt them for the MIG task. We collect challenging examples from Objects365~\cite{shao2019objects365} for Common Object Grounding and Region Locating. We use examples that exhibits significant movement from GOT-10k~\cite{huang2019got} for Object Tracking. For Multi-View Grounding, we utilize the rich annotations from Ego4D~\cite{grauman2022ego4d}. The Static Difference task is sourced from MagicBrush~\cite{zhang2024magicbrush}. We combine multiple examples from GranD~\cite{rasheed2024glamm} to form a group for Group Grounding. Additionally, for tasks such as Reasoning Grounding, Correspondence, Referring Grounding, and Robust Diff Grounding, which lack suitable existing datasets, we collect both web images and manually captured photos, and annotate them with well-educated annotators who are thoroughly trained and fully understand these tasks. 

All instances are reviewed by two different human annotators to guarantee the quality of \ourbench. This includes removing instances with incorrect annotations, ensuring that the questions are answerable, filtering out overly simplistic questions, and refining ambiguous queries. We also invite five volunteers to answer the questions to evaluate human performance on this benchmark (detailed in Appendix \ref{appendix:human_eval}). As shown in Table~\ref{tab:mig-bench}, the average accuracy of human responses is 97.18\%, indicating that the task is easy for humans and further demonstrating the high quality of \ourbench.

Unlike existing benchmarks, \ourbench introduces the grounding task in a multi-image setting, thereby addressing the gap in previous benchmarks that are unable to measure the free-form MIG capabilities of MLLMs. A detailed comparison with other benchmarks is provided in Appendix~\ref{appendix:Benchmark Comparison}.

\section{Experiments}

\subsection{Implementation Details}
Migician undergoes development based on the Qwen2-VL-7B~\cite{wang2024qwen2} foundation model with a global batch size of 48, a total of 25,000 steps for the two-stage training procedure, and a learning rate of 5e-6, using 8×A100-80G GPUs. For the evaluation in our proposed \ourbench, we use the conventional metric $\text{Acc}_{0.5}$ in referring expression comprehension~\cite{kazemzadeh2014referitgame}. This metric measures the accuracy of object localization, defining a prediction as correct if the Intersection over Union (IoU) with the ground truth is greater than 0.5.

\subsection{Results on MIG-Bench}

As shown in Table~\ref{tab:mig-bench}, Migican achieves the state-of-the-art performance across all tasks on MIG-bench, with an average improvement of 24.94\% compared to the second-best model, Qwen2-VL-72B (38.88\%), despite having significantly fewer parameters. Note that there is a substantial gap between human performance and that of all MLLMs across all tasks, indicating that MLLMs have significant potential for improvement in free-form MIG. In particular, for 7B-scale models, even advanced multi-image models like InternVL2-8B and Qwen2-VL-7B struggle to perform, particularly in tasks such as multi-view grounding, region locating, and correspondence.

For models equipped with grounding capabilities, such as mPLUG-Owl3, InternVL2 series, and Qwen2-VL series, they demonstrate an advantage over other baselines.
Furthermore, the proposed single-image CoT method~(\textit{+CoT}) effectively integrates the grounding and multi-image understanding capabilities of the MLLMs where different abilities assist each other in different reasoning steps, achieving comprehensive improvements on multi-image grounding tasks. Moreover, this approach proves effective for all the aforementioned models.

\begin{table*}[t]
\centering
\begin{minipage}{0.59\textwidth}
\centering
\resizebox{\textwidth}{!}{%
\begin{tabular}{c|cccccc}
\toprule
\textbf{Model} & \textbf{MuirBench} & \textbf{BLINK val} & \textbf{MIBench} & \textbf{Mantis\_eval} & \textbf{MMIU} & \textbf{AVE}\\ 
\midrule 
\rowcolor[HTML]{E6E6E6}
\multicolumn{7}{c}{\textbf{Closed-Source Model}}\\
\midrule 
GPT-4o & 62.31 & 60.04 & 71.88 & 62.67 & 55.7 & 62.52\\ 
Gemini-Pro & 49.35 & 45.16 & — & — & 53.4 & 49.30\\ 
\midrule 
\rowcolor[HTML]{E6E6E6}
\multicolumn{7}{c}{\textbf{Open-Source Model}}\\
\midrule
LLaVA-1.5 & 23.46 & 37.13 & 26.83 & 31.34 & 19.20 & 27.59\\
CogVLM & 20.85 & 41.54 & — & 45.16 & 23.57  & 32.78\\
Idefics2-8B & 26.08 & — & 46.39 & 48.85 & 27.80 & 37.28\\
mPLUG-Owl3 & 39.67 & 50.30 & 56.66 & 63.10 & 21.72 & 46.29\\  
InternVL2-8B & 48.70 & 50.57 & 52.91 & 60.37 & 42.00 & 50.05\\ 
Mantis & \textit{\underline{44.50}} & 49.05 & 45.09 & 57.14 & 45.60 & 48.28\\
LLaVA-OV-7B & 41.80 & 48.20 & \textit{\underline{71.29}} & 64.20 &  44.46 & 53.99\\ 
MiniCPM-V 2.6 & 42.65 & 51.45 & 71.09 & \textit{\underline{69.12}} & 50.19 & 56.90\\ 
Qwen2-VL-7B & 42.04 & \textbf{52.35} & 68.06 & \textbf{70.97} & \textit{\underline{54.36}} & \textit{\underline{57.56}}\\
\midrule
\rowcolor[HTML]{E6E6E6}  
Migician & \textbf{53.69} & \textit{\underline{51.53}} & \textbf{71.42} & \textit{\underline{69.12}} & \textbf{60.32}& \textbf{61.51}\\ 
\bottomrule
\end{tabular}%
}
\caption{Performance comparison on various multi-image understanding benchmarks. The
highest score is highlighted in bold and the second highest score is \textit{underlined} for all open-source models.}
\label{tab:general_bench}
\end{minipage}
\hfill 
\begin{minipage}{0.39\textwidth}
\centering
\resizebox{\textwidth}{!}{%
\begin{tabular}{c|ccc}
\toprule
\textbf{Model} & \textbf{MME} & \textbf{MMBench} & \textbf{V* Bench}\\ 
\midrule 
\rowcolor[HTML]{E6E6E6}
\multicolumn{4}{c}{\textbf{Closed-Source Model}}\\
\midrule 
GPT-4V & 1926.6 & \textit{\underline{81.0}} & 54.97 \\ 
Gemini-Pro & 2148.9 & 73.6 & 48.16 \\ 
Claude-3.5 & 1920.0 & 79.7 & — \\
\rowcolor[HTML]{E6E6E6}  
\midrule 
\rowcolor[HTML]{E6E6E6}
\multicolumn{4}{c}{\textbf{Open-Source Model}}\\
\midrule 
LLaVA-1.5 & 1510.7 & 64.3 & 48.68 \\ 
InternVL2-8B & \textit{\underline{2210.3}} & \textbf{81.7} & 43.07\\
MiniCPM-V 2.6 & 2024.6 & 77.2 & 52.67 \\
SEAL & 1128.9 & 33.1 & \textbf{75.39} \\ 
LLaVA-OV-7B & 1998.0 & 80.9 & — \\
Mantis & 1806.4 & 75.7 & — \\
\midrule
\rowcolor[HTML]{E6E6E6}  
Migician & \textbf{2244.7} & 80.0 & \textit{\underline{72.30}} \\ 
\bottomrule
\end{tabular}%
}
\caption{The performance of models on various single-image benchmarks, where Migician consistently exhibits strong capabilities.}
\label{tab:single_image_res}
\end{minipage}
\end{table*}

\begin{table*}[t]
\centering
\begin{minipage}{0.59\textwidth}
\centering
\resizebox{\textwidth}{!}
{
\begin{tabular}{c|ccc|ccc|cc|c}
\toprule
\multicolumn{1}{c|}{\multirow{2}{*}{\textbf{Model}}} & 
\multicolumn{3}{c|}{\textbf{RefCOCO}} & 
\multicolumn{3}{c|}{\textbf{RefCOCO+}} & 
\multicolumn{2}{c|}{\textbf{RefCOCOg}} & 
\multirow{2}{*}{\textbf{AVE}} \\ 
\cmidrule{2-9}
& val & testA & testB & val & testA & testB & val & test & \\
\midrule
VisionLLM v2~\cite{wu2024visionllm} & 79.20 & 82.30 & 77.00 & 68.90 & 75.80 & 61.80 & 73.30 & 74.80 & 74.14\\
Shikra~\cite{chen2023shikra} & 87.00 & 90.60 & 80.20 & 81.60 & 87.40 & 72.10 & 82.30 & 82.20  & 82.97\\
InternVL2-8B~\cite{cai2024internlm2} & 87.10 & 91.10 & 80.70 & 79.80 & 87.90 & 71.40 & 82.70 & 82.70 & 82.94\\ 
GroundingGPT~\cite{li2024groundinggpt} & 88.02 & 91.55 & 82.47 & 81.61 & 87.18 & 73.18 & 81.67 & 81.99 & 83.57\\
Griffon v2~\cite{zhan2024griffon} &89.6& 91.80 & 86.50 & 81.90 & 85.50 & 76.20 & 85.00 & 86.00 & 85.30\\
InternVL2-8B~\cite{cai2024internlm2} & 87.10  & 91.10 &  80.70 &  79.80 &  87.90 &  71.40 &  82.70 &  82.70 &  82.94 \\
Qwen2-VL-7B~\cite{wang2024qwen2} & \textbf{91.70} &  \textbf{93.60} &  \textbf{87.30} &  85.80 &  90.50 &  79.50 &  87.30 &  \textbf{87.80} &  87.96\\
\midrule
\rowcolor[HTML]{E6E6E6}  
Migician & 91.62 & 93.49 & 87.22& \textbf{86.13} & \textbf{91.06} & \textbf{79.93} & \textbf{88.06} & \textbf{87.80} & \textbf{88.16}\\ 
\bottomrule
\end{tabular}
}
\caption{The performance of different competitive single image grounding models on Refcoco, Refcoco+ and Refcocog benchmarks. Continual grounding training in the multi-image scenario further enhances the overall grounding ability of the model. Migician achieves top performance among all grounding models.}
\label{tab:refcoco}
\end{minipage}
\hfill 
\begin{minipage}{0.385\textwidth}
\centering
\resizebox{\textwidth}{!}{%
\begin{tabular}{l|ccc}
\toprule
\textbf{Models} & \textbf{Spontaneous} & \textbf{Referential} & \textbf{AVE}\\ 
\midrule
mPLUG-Owl3 & 19.96 & 9.08 & 13.04\\
mPLUG-Owl3$_{+mCoT}$ & 23.78 & 14.10 & 17.62 \\
mPLUG-Owl3$_{+CoT}$ & 26.73 & 26.43 & 26.54 \\
\midrule
InternVL2-8B & 13.29 & 17.10 & 15.71\\
InternVL2-8B$_{+mCoT}$ & 23.78 & 21.99 & 22.64 \\
InternVL2-8B$_{+CoT}$ & 31.52 & 37.57 & 35.37 \\
\midrule
Qwen2-VL-7B & 19.96 & 28.67 & 28.61\\
Qwen2-VL-7B$_{+mCoT}$ & 41.83 & 26.23 & 31.90 \\
Qwen2-VL-7B$_{+CoT}$ & 42.59 & 44.34 & 43.70 \\
\bottomrule
\end{tabular}
}
\caption{The comparison among different CoT variants. We compare three representative MLLMs among direct reference, single-image CoT~(\textit{+CoT}), multi-image CoT~(\textit{+mCoT}) as described in Section \ref{sec:mcot}.}
\label{tab:mCoT}
\end{minipage}
\end{table*}

\begin{table*}[t]
\centering
\resizebox{\textwidth}{!}{
\begin{tabular}{l|ccccc|c}
\toprule
\multirow{2}{*}{\textbf{Setting}} & \multicolumn{5}{c|}{\textbf{Multi-image General Benchmarks}} &  \multirow{2}{*}{\textbf{MIG}} \\ 
\cmidrule(lr){2-6} 
& \textbf{MuirBench} & \textbf{BLINK} & \textbf{MIBench} & \textbf{Mantis} & \textbf{MMIU} & \\ 

\midrule
Base & 42.04 & 52.35 & 68.06 & 70.97 &  54.36 & 28.62\\
\midrule
Full data & 53.77 & 51.27 & 71.76 & 66.36 & 53.31 & 62.79 \\
-w/o grounding & 44.54$_{(-9.23)}$ & 51.32$_{(+0.42)}$ & 71.68$_{(-0.08)}$ & 67.74$_{(+1.38)}$ & 52.12$_{(-1.19)}$ & 22.43$_{(-40.36)}$ \\
-w/o general & 53.62$_{(-0.15)}$ & 49.25$_{(-2.02)}$ & 65.22$_{(-6.54)}$ & 64.52$_{(-1.84)}$ & 48.61$_{(-4.70)}$ & 62.21$_{(-0.58)}$ \\

\bottomrule
\end{tabular}
}
\caption{The ablation study on the contribution of different data subsets.}
\label{tab:data_ablation}
\end{table*}

\begin{table}[t]
\centering
\resizebox{\columnwidth}{!}{
\begin{tabular}{c|ccc}
\toprule
\textbf{Model} & \textbf{Easy} & \textbf{Medium} & \textbf{Hard}\\ 
\midrule 
\# Instances	& 2471 	& 1430 	& 395 \\
\midrule 
InternVL2-8B	& 44.69 	& 13.92 	& 1.77 \\
Qwen2-VL-7B	& 30.31 	& 22.24 	& 0.00 \\
InternVL2-8B$_{+CoT}$	& 67.10 	& 7.06 	& 0.25 \\
Qwen2-VL-7B$_{+CoT}$	& 71.02 	& 10.70 	& 0.76 \\
Migician	& \textbf{76.00} 	& \textbf{52.10} 	& \textbf{29.37} \\

\bottomrule
\end{tabular}
}
\caption{Performance comparison across varying difficulty levels on \ourbench.}
\label{tab:MIG-Complexity}
\end{table}

\subsection{Results on Multi-Image Understanding Benchmarks}

As shown in Table~\ref{tab:general_bench},  Migician not only establishes its multi-image grounding ability, but also remarkably stimulates its general multi-image understanding ability. In particular, \ours achieves the best average results on the multi-image understanding benchmarks. It surpasses the second-best model~(Mantis) on MuirBench by 9.19\%, achieving SOTA performance on both MMIU and MIBench. We attribute this to the training on a mixture of  multi-image understanding and grounding data, which indicates that our proposed \ourtrain dataset can effectively enhance general multi-image comprehension.

\subsection{Results on Single-Image Benchmarks}

Table~\ref{tab:single_image_res} lists the empirical results on typical single-image understanding benchmarks including MME~\cite{fu2024mmecomprehensiveevaluationbenchmark} and MMBench~\cite{liu2024mmbench}, suggesting that \ours retains strong single-image understanding capacities. Specifically, on the MME benchmark, \ours surpasses notable MLLMs like InternVL2-8B and MiniCPM, while exhibiting equally strong performance with close-source models on MMBench. Notably, contrasting with specialized multi-image models such as LLaVA-OV and Mantis, whose single image ability has largely degenerated, \ours poses significant advantage over them, achieving comprehensive capability maintenance. Our model also maintains strong performance on single-image REC tasks, as shown in Table \ref{tab:refcoco}.

Furthermore, we find that the MIG ability of \ours can be leveraged to address the task of finding visual details in high-resolution images, such as V*Bench~\cite{wu2024v}. Specifically, we split a single high-resolution image in V*Bench into multiple sub-images and transform the problem into a MIG task (detailed in Appendix~\ref{appendix:v_search}). Results show that \ours can generalize well to this out-of-distribution setting, performing on par with the specialized visual searching system SEAL~\cite{wu2024v}.

\section{Analysis}
\subsection{Effects of Different CoT Strategies}
\label{sec:mcot}
The CoT framework in Section~\ref{sec:cot}, after obtaining a referring expression, has the MLLM perform grounding in each image in a polling manner (denoted as single-image CoT), which incurs significant inference overhead. Here, we explore multi-image CoT, where the MLLM directly performs grounding across all images based on the obtained referring expression. As shown in Table \ref{tab:mCoT}, multi-image CoT achieves some effectiveness but it still falls significantly behind single-image CoT. In contrast, our proposed \ours is able to perform end-to-end reasoning under the multi-image setting, offering significant advantages in both efficiency and effectiveness.

\subsection{Effects of Different Data on Multi-Image Understanding}

As observed in Table \ref{tab:general_bench}, \ours demonstrates an improvement in the multi-image understanding. We further conduct an ablation study to analyze the effects of different data subsets.
Specifically, we train two models with either \ourtrain or multi-image understanding data removed from the original training set. 
The results in Table \ref{tab:data_ablation} reveal that grounding data generally contributes to multi-image understanding. In 4 out of 5 benchmarks, the full dataset achieves the highest performance compared to models trained with any subset of data removed. In contrast, directly fine-tuning with only general data does not consistently lead to a performance boost. However, when combined with fine-grained grounding data, the model experiences a notable improvement.

\subsection{Performance across Difficulty Levels}

To comprehensively assess \ours across varying scenario complexities in \ourbench, we have established a three-tier difficulty classification (Easy/Medium/Hard) through joint consideration of three key factors: (1) the number of input images, (2) the accuracy rates of four representative baselines (Qwen2-VL, InternVL2, and their CoT variants), and (3) the average IoU improvement when applying CoT. Specifically, an instance is classified as \textbf{Easy} if either (a) more than two models achieve correct answers with fewer than four input images, or (b) the CoT-enhanced models demonstrate an IoU improvement exceeding 0.15. Conversely, instances are deemed \textbf{Hard} when no more than one model succeeds despite processing over four input images. All remaining cases that fall between these thresholds are categorized as \textbf{Medium} difficulty.

Through this way, we can evaluate model capabilities across diverse scenario complexities. As shown in Table~\ref{tab:MIG-Complexity}, Migician demonstrates remarkable performance advantages across all difficulty levels while exhibiting particularly strong capabilities in challenging scenarios. Noticeably, the performance gap widens significantly in medium and hard difficulty settings as Migician achieves approximately 30\% greater accuracy compared to the baseline models. This pronounced advantage in complex scenarios highlights Migician's superior capacity for free-form MIG.

\section{Conclusion}
In this work, we explore the task of multi-image grounding and propose \ours, the first MLLM to overcome the barriers between fine-grained visual grounding and multi-image inputs. With our proposed large-scale \ourtrain dataset, \ours seamlessly integrates grounding across multiple images, enabling free-form multi-image grounding. To further advance research in this area, we introduce \ourbench, a comprehensive benchmark for evaluating the multi-image grounding capabilities of MLLMs. Experimental results demonstrate that our model significantly outperforms existing methods. We hope this work will inspire further developments in multi-image grounding and contribute to the creation of more versatile multimodal models in the future.

\section*{Limitation}
Despite our comprehensive discussion of the MIG challenge, several limitations remain. First, due to computational constraints, we have not verified the effectiveness of our training methods on larger 70B-scale models. Second, the current model can still produce hallucinated outputs sometimes as conventional MLLMs. 
Lastly, our training methods and benchmark construction mainly focus on the REC task. Although Migician possesses decent REG capacity, this topic is still insufficiently discussed.

\section*{Acknowledgement}
This work is supported by the Fundamental Research Funds for the Central Universities of China under Grant 2024JBGP008 and the National Natural Science Foundation of China (No. 62406018). 

We extend our heartfelt gratitude to the dedicated human volunteers, Mai Sun, Pujian Zhan, Xingyu Zhang, Binhao Liu, and Huiting Pei, for their tireless efforts in human-level performance evaluation, for which we extend our wholehearted appreciation. 

{
    \small
    \bibliographystyle{ieeenat_fullname}
    \bibliography{main}
}

\clearpage

\appendix

\section{Benchmark Tasks Definition}
\label{appendix:Benchmark Definition}

\begin{table*}[t]
\resizebox{\textwidth}{!}{
\begin{tabular}{l|cccccccc}
\toprule
\textbf{Dataset} & \textbf{Images} & \textbf{Ave-I} & \textbf{Max-I} & \textbf{Multi-Image} & \textbf{Multi-Task} & \textbf{Instance-Labeled} & \textbf{Instances} & \textbf{Reference}\\ 
\midrule
Q-Bench & 3489 & 2.0 & 2.0 & \textcolor{myred}{\hcmark} & \textcolor{mygreen}{\cmark} & \textcolor{myred}{\ding{55}} & \textcolor{myred}{\ding{55}} & \textcolor{myred}{\ding{55}}\\
Mantis-Eval & 542 & 2.5 & 5.0 & \textcolor{mygreen}{\cmark} & \textcolor{myred}{\ding{55}} & \textcolor{myred}{\ding{55}} & \textcolor{myred}{\ding{55}} & \textcolor{myred}{\ding{55}} \\
BLINK & 3612 & 1.9 & 4.0 & \textcolor{mygreen}{\cmark} & \textcolor{mygreen}{\cmark} & \textcolor{myred}{\ding{55}} & \textcolor{myred}{\ding{55}} & \textcolor{myred}{\ding{55}} \\
MIRB & 3497 & 3.8 & 42.0 & \textcolor{mygreen}{\cmark} & \textcolor{mygreen}{\cmark} & \textcolor{myred}{\ding{55}} & \textcolor{myred}{\ding{55}} & \textcolor{myred}{\ding{55}} \\
\midrule
Refcoco/g/+ & 3900 & 1.0 & 1.0 & \textcolor{myred}{\ding{55}} & \textcolor{myred}{\ding{55}} & \textcolor{mygreen}{\cmark} & 7596 & T \\
HC-Refcoco/+/g & 1521 & 1.0 & 1.0 & \textcolor{myred}{\ding{55}} & \textcolor{myred}{\ding{55}} & \textcolor{mygreen}{\cmark} & 3754 & T \\
GigaGround & 3775 & 1.0 & 1.0 & \textcolor{myred}{\ding{55}} & \textcolor{myred}{\ding{55}} & \textcolor{mygreen}{\cmark} & 61353 & T\\
MC-Bench & 3345 & 2.0 & 2.0 & \textcolor{myred}{\hcmark} & \textcolor{mygreen}{\cmark} & \textcolor{mygreen}{\cmark} & 3202 & T \\
\midrule
MIG-Bench & 5887 & 3.8 & 6.0 & \textcolor{mygreen}{\cmark} & \textcolor{mygreen}{\cmark} & \textcolor{mygreen}{\cmark} & 4295 & T/I/T+I \\
\bottomrule
\end{tabular}
}
\caption{Comparison of MIG-Bench with other benchmarks.}
\label{tab:benchmark_comparison}
\end{table*}

\subsection{Spontaneous Grounding}
Our benchmark evaluates spontaneous grounding through three distinct tasks below, which aim at assessing model's ability to autonomously discover insidious connections across various images and then accurately locate the target.
\paragraph{Spot the Difference} Similar to the spot-the-difference puzzle, given two similar images with a single difference, the model is instructed to recognize and ground this difference in the second image, requiring simultaneous and keen perception of both images. 
\paragraph{Common Object Grounding} It refers to automatically recognizing and grounding the common object appearing in all images within an image group, which shares one definite common object in our benchmark.
\paragraph{Robust Image Difference Grounding} Models must focus on the primary difference between two images captured from slightly different perspectives, ignoring other minor variations caused by shifts in the viewpoint. The incorporation of view changes presents a greater challenge for the model and better reflects real-world scenarios, where variations in perspective are inevitable.
\subsection{Reference Grounding}
\paragraph{Textual Reference Query} This challenge, which mainly includes \textbf{Group Grounding}, tests a model's ability to link a textual reference to the target object within its corresponding image among an image group. Given a set of images and one textual query, the model must firstly identify the correct image then accurately ground the target object within it, additionally incorporating image-level locating compared with conventional grounding task.

\paragraph{Visual Reference Query} These tasks focuses on effectively utilizing visual reference information and incorporate it into the locating process.

(1) \textbf{Visual Referring Grounding.} In this task, a pair of images is provided—a source image with a clear object and a target image containing multiple elements. The model must perceive the referenced object of the source image and then locate it in the target image.

(2) \textbf{Region Locating.} Models are tasked with identifying multiple region images within a source image, which often requires perceptive and discerning observation as the model may encounter person recognition, similar object distinguishing, tiny item searching and etc.

(3) \textbf{Object Tracking.} This task involves tracking a target object across a sequence of video frames. The object is highlighted with a red bounding box in the first image, and the model must keep track of it throughout the entire sequence.

(4) \textbf{Multi-view Grounding.} Here, the model must locate the same target across multiple images taken from distinctive viewpoints, for instance egocentric view and exocentric view.
\paragraph{Visual+Textual Reference Query} These tasks combine information from both modalities to assess cross-modal reasoning\&grounding abilities.

(1) \textbf{Correspondence.} The model must ground semantically or functionally similar regions within the target image. This finer-grained task focuses on object regions rather than whole objects, demanding an in-depth understanding of visual semantics. 

(2) \textbf{Reasoning.} This task requires the model to perform reasoning-based grounding by integrating cross-modality information. Several examples are shown in Figure~\ref{fig:MIGtasks}.

Our MIG-Bench offers a rich, multi-faceted evaluation across various real-world scenarios and domains, extending beyond simple image pairs to include longer and more complex image contexts. By ensuring that each task is well-defined and unambiguous, we facilitate objective and definitive assessments.

\section{Comparison of MIG-Bench with Other Benchmarks}
\label{appendix:Benchmark Comparison}

A detailed comparison with other benchmarks are provided in Table~\ref{tab:benchmark_comparison}. Current evaluations for MLLMs primarily focus on single-image perception, understanding, reasoning, or grounding (e.g., MME, MMBench, Refcoco), leaving the multi-image scenario largely unexplored. While benchmarks such as Mantis-eval, BLINK, and MIRB are representative of multi-image evaluation, they concentrate on image-level and general understanding of multiple images, failing to comprehensively assess a model's fine-grained grounding skills in the multi-image scenario.

Unlike traditional grounding benchmarks or multi-image benchmarks, MIG-Bench introduces the grounding task into a multi-image scenario, covering a series of 10 distinct tasks. The most relevant benchmark to our work is MC-Bench, a contemporaneous study. MC-Bench evaluates the multi-context grounding capabilities of MLLMs by asking them to accurately locate the corresponding object based on a text prompt in the correct image from a given image pair. However, it has limitations in terms of the fixed number of input images and the restricted forms of queries. In contrast, the proposed MIG-Bench offers more flexible task formats, focusing on evaluating models' abilities in free-form multi-image understanding. 

\section{Single-Image CoT Failure Patterns}
\label{appendix:failure}
\begin{figure*}[h] 
    \centering
    \includegraphics[width=0.92\textwidth]{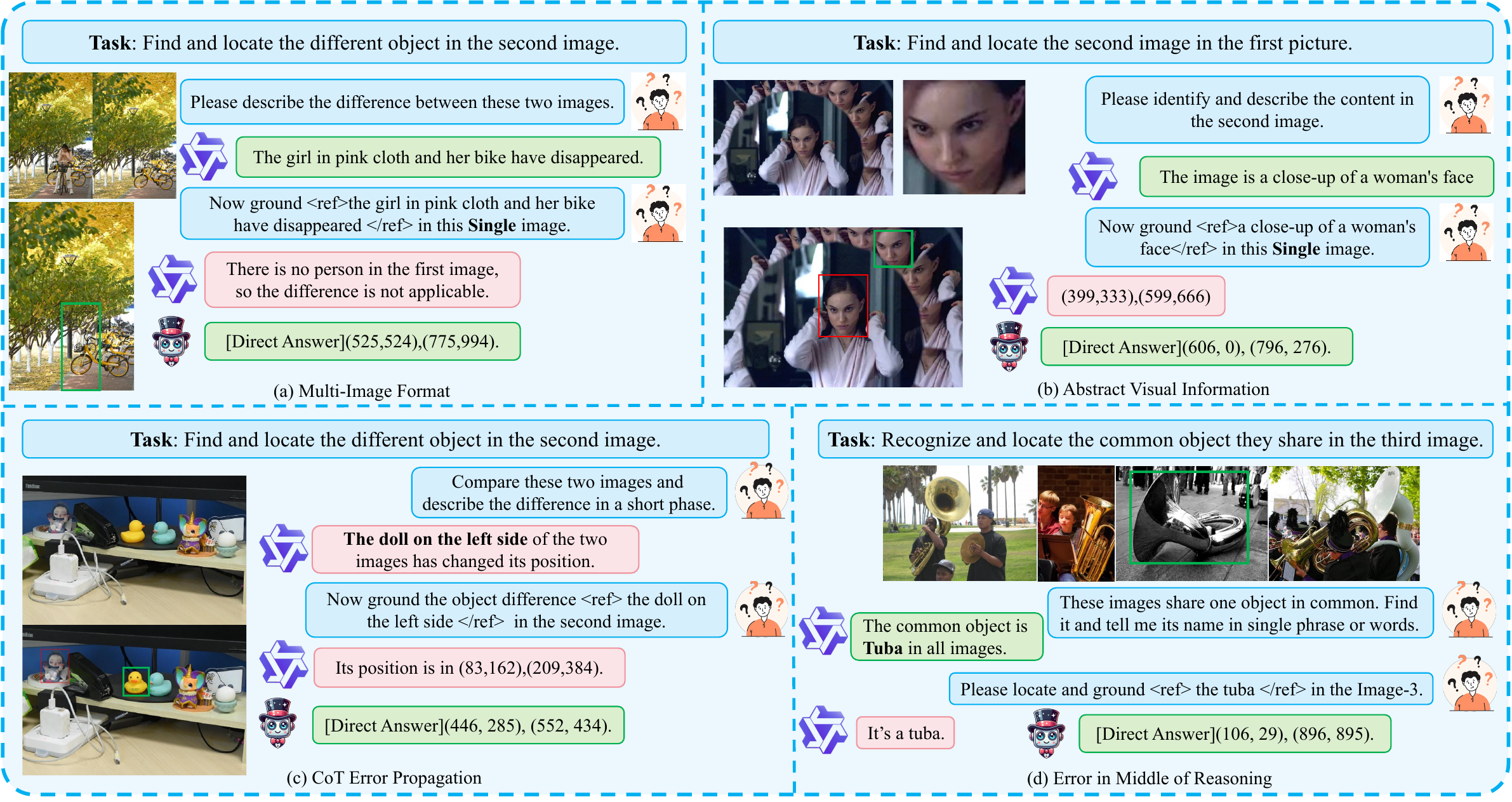} 
    \caption{Above are the four representative failure patterns of the single-image CoT. From left to right, top to bottom, they are (a) special multi-image format,  (b) abstract visual information, (c) CoT error propagation, (d) step-2 inference error.} 
    \label{fig:failure} 
\end{figure*}

We have analyzed more failure patterns of the CoT framework in Figure~\ref{fig:failure}, categorized into perceptual and reasoning flaws. 

For the former, the framework falls short when multiple images are organized in a manner where only integrating all their visual information could tackle MIG (i.e. finding the location of missing people in the second image), or when the textual content could not sufficiently represent the visual information. 

Regarding reasoning errors, inaccuracies can arise at various stages of the reasoning process, undermining the framework’s overall accuracy and effectiveness.

These failure patterns highlight the significant limitations of simply integrating the different capabilities of current models through a simple CoT framework, underscoring the need for an end-to-end model capable of directly performing the MIG task.

\section{\ourtrain Data Curation Details}
\subsection{Transforming Existing Data}
\label{appendix:stage1data}
\paragraph{Static Diff}
Describing the differences among two nearly identical pictures is a well discussed topic, yet previous attempts capture the differences through textual phrases, failing to precisely recognize their locations. After a comprehensive survey on this area, we have collected high-quality and fully labeled image difference data from various existing datasets: Spot-the-diff~\cite{jhamtani2018learning}, Img-diff~\cite{jiao2024img}, MagicBrush~\cite{zhang2024magicbrush} and CLEVR-change~\cite{park2019robust}. 

By these collected datasets inherently contain much noise in them, for instance, inaccurate difference caption, incorrect bounding box coordinates and etc. We filter the inaccurate bbox labels from Spot-the-Diff and only preserve the correct ones and their difference captions. Additionally, we conduct down-sampling on Img-Diff due to its diffusion generation~\cite{rombach2022high} based nature and consequent inaccuracies.

During the construction process, we ensure the diversity of the content by (1) incorporating numerous prompt formats generated by GPT-4 and improving the instruction-following ability of the model, (2) constructing CoT process to assist the model gradually and progressively reaching the final answer by fully utilizing the annotation available in the dataset.

\begin{table*}[t]
\resizebox{\textwidth}{!}{%
\begin{tabular}{c|cccccc}
\toprule
\textbf{Training Methods} & \textbf{Referring} & \textbf{Object Tracking} & \textbf{Group Grounding} & \textbf{Region} & \textbf{Static Diff} & \textbf{Common Object} \\ 

\midrule
Base & 23.23 & 20.73 & 58.52 & 25.95 & 27.84 & 19.36 \\
\midrule
Multi-Task Learning & 60.00 & 61.65 & 62.28 & 57.95 & 55.68 & 81.37 \\
Separate Learning & 69.70 & 74.55 & 63.13 & 65.42 & 68.94 & 79.53 \\
Model Merging & 60.61 & 50.00 & 64.53 & 18.95 & 29.92 & 65.44\\
\bottomrule
\end{tabular}%
}
\caption{Comparison between different training methods. We compare the learning efficiency between multi-task learning, separate learning and merging all these task-specialized modes. We mainly focus on the in-domain tasks that M-Grounding dataset covers.}
\label{tab:separate}
\end{table*}

\begin{table*}[t]
\resizebox{\textwidth}{!}{%
\begin{tabular}{c|c|cccc}
\toprule
\textbf{Models} & \textbf{Settings} & \textbf{Common Object} & \textbf{Multi-view Grounding} & \textbf{Object Tracking} & \textbf{Region Locating}\\ 

\midrule
— & Random Guess & 26.47 & 1.04 & 2.13 & 0.00 \\
\midrule
Qwen2-VL-7B & Polling & 19.96 & 11.83 & 20.73 & 25.95\\
Qwen2-VL-7B & All & 19.36 & 6.60 & 13.09 & 11.80\\
Qwen2-VL-7B & Polling+mCoT & 53.80 & 14.24 & 21.09 & 20.20\\
Qwen2-VL-7B & All+mCoT & 45.71 & 9.38 & 17.55 & 15.54\\
Migician & Polling & 81.99 & 44.44 & 61.09 & 59.65\\
Migician & All & 72.43 & 43.06 & 58.55 & 34.91\\

\bottomrule
\end{tabular}%
}
\caption{Comparison of different answering forms. For random guess, we set the default answer as \textit{(0,0),(999,999)}.}
\label{tab:polling}
\end{table*}

\paragraph{Common Object Grounding}
Grounding the primary common object across multiple images is a challenging task for models. It requires them to simultaneously perceive multiple images, isolate the common object, and then accurately ground the target within each image.

In our approach, we leverage diverse data sources, including ImageNet~\cite{deng2009imagenet}, COCO~\cite{lin2014microsoft}, and Object365, which are rich in extensive annotations. To organize the images, we group those containing the same object based on object labels, applying a threshold to filter out objects that occupy too small a proportion of the image. This threshold, determined empirically, effectively mitigates ambiguity when multiple candidate objects could be considered the common object, thereby producing clear and definitive training examples. We further reduce ambiguity by removing classes that often co-occur with other object, for instance keyboard, knife, couch, dinning table and etc.

\paragraph{Object Tracking}
The multi-image setting is well-suited for the object tracking task, which involves both temporal and spatial grounding. To build our dataset, we have carefully selected large-scale, well-annotated datasets including TrackingNet~\cite{muller2018trackingnet}, LaSOT~\cite{fan2019lasot}, GOT-10K~\cite{huang2019got}, and MOT-2017~\cite{milan2016mot16}. During dataset construction, we extract 4-6 images from each original sequence at specific intervals, preserving key features while maintaining efficiency. Additionally, we introduce a small proportion of temporal order judgment data from continuous video frames to strengthen the model's temporal reasoning capabilities.

\paragraph{Referring Grounding}
This training data simulates the process of locating an object from a source image in a target image. We primarily use the ImageNet-2012 dataset to create image pairs, where the source object is fully visible in the first image, and the target object is hidden in the second. Empirically, this design generates challenging training examples that require the model to first recognize the source object and then search for it in the target image.

\paragraph{Group Grounding}
Conventional visual grounding is typically limited to single-image contexts. However, in real-world scenarios, it is often necessary to recognize a target object within a cluttered collection of images. Group Grounding addresses this limitation by enabling the identification of the target among a group of images, thereby enhancing the versatility of traditional grounding methods.

For the construction of Group Grounding training data, we leverage the large-scale GranD rec and reg conversation dataset~\cite{rasheed2024glamm}, which contains 3 million examples. After filtering out noisy data and grouping 3-5 images per set, we curate a high-quality collection of 120k instances for stage-1 training. This dataset effectively enhances the model's ability to perform image co-reference~\cite{jiang2024mantis}, image-level grounding, and instance-level localization.

\paragraph{Region Locating}
Region locating involves slicing an image into several semantically rich regions and identifying the precise locations of these regions within the source image. To extract meaningful regions, we use the Objects365 dataset, selecting labeled bounding box areas as the regions of interest. To further enhance the quality of the regions, we apply a series of filtering criteria:
(1) Content Richness: We select images with more than 10 bounding box annotations to avoid overly simple or plain cases.
(2) Aspect Ratio: We retain regions with an aspect ratio between 0.5 and 2 to exclude excessively narrow bounding boxes that may be difficult for the model to handle.
(3) Size: We ensure that the region-to-image ratio lies between 0.2 and 0.49, with an absolute pixel count above 2,000, to exclude tiny and obscure regions that may lack sufficient detail.
Notably, due to our carefully designed filtering mechanism and the inherent characteristics of the task, the resulting training data predominantly includes cases of person recognition, distinguishing between similar objects (i.e. chairs, bowls, cars and etc), and recognizing tiny details—tasks that are non-trivial even for humans.

\begin{table*}[t] 
\centering

\resizebox{\textwidth}{!}{
    \begin{tabular}{l|ccc | ccccccc|c}
    \toprule
    \multirow{4}{*}{\textbf{Models}} & \multicolumn{3}{|c}{\textbf{Spontaneous Grounding }} & \multicolumn{7}{|c|}{\textbf{ Referential Grounding }} & \multirow{4}{*}{\textbf{AVE}} \\ 
    \cmidrule(lr){2-4} \cmidrule(lr){5-11} 
    
    \multicolumn{1}{c}{} & \multicolumn{2}{|c|}{\textbf{Difference}} & \textbf{Similarity} &  \multicolumn{4}{c|}{\textbf{Visual Reference}} & \textbf{Textual} & \multicolumn{2}{|c|}{\textbf{Visual+Textual}}& \\ 
    \cmidrule(lr){2-3} \cmidrule(lr){4-4} \cmidrule(lr){5-8} \cmidrule(lr){9-9} \cmidrule(lr){10-11}
    
    \multicolumn{1}{c|}{} & \textbf{Static} & \textbf{Robust} & \multicolumn{1}{|c|}{\textbf{Common}}& \textbf{OT} & \textbf{MV} & \textbf{Region} & \textbf{Refer} &  \multicolumn{1}{|c|}{\textbf{GG}} & \textbf{Reason} & \multicolumn{1}{c|}{\textbf{Co-Re}} & \\
    
    \midrule

    \rowcolor[HTML]{ACE8A1}
    \multicolumn{12}{c}{\textbf{70B Scale Models}}\\
    \midrule

    LLaVA-OV-72B & 13.26 & 5.34 & 26.84 & 12.91 & 7.64 & 2.14 & 17.83 & 21.60 & 11.88 & 8.55 & 13.65 \\
    InternVL2-76B & 15.91 & 10.64 & 36.40 & 30.73 & 20.83 & 5.74 & 46.46 & 41.28 & 32.67 & 26.50 & 26.72 \\
    Qwen2-VL-72B & \textbf{46.12} & 46.81 & 64.46 & 26.73 & 22.57 & 18.62 & 33.33 & 62.53 & 50.50 & 17.09 & 38.88 \\

    LLaVA-OV-72B$_{+CoT}$ & 20.27 & 21.28 & 52.57 & 44.36 & 20.83 & 25.60 & 37.37 & 35.07 & 31.68 & 28.21 & 31.72 \\
    InternVL2-76B$_{+CoT}$ & 16.86 & 6.38 & \textbf{70.34} & \textbf{70.55} & 33.33 & 27.27 & 68.69 & 57.31 & 52.48 & 23.08 & 42.63 \\
    Qwen2-VL-72B$_{+CoT}$ & 33.33 & \textbf{47.87} & 69.24 & 70.18 & \textbf{60.42} & \textbf{51.04} & \textbf{78.79} & \textbf{70.74} & \textbf{70.30} & \textbf{35.04} & \textbf{58.70} \\
    
    \bottomrule
    \end{tabular}
}
\caption{Performance Comparison of 70B scale models equipped with CoT.}
\label{tab:70B}
\end{table*}

\begin{table}[t]
\centering
\resizebox{\columnwidth}{!}{
\begin{tabular}{c|cc}
\toprule
\textbf{Type} & \textbf{Source} & \textbf{Ratio} \\ 
\midrule 
\rowcolor[HTML]{E6E6E6} 
\multicolumn{3}{c}{\textbf{Stage-1}} \\ 

\midrule 
S-Understanding & LLaVA-OV-data & 17\% \\ 
S-Grounding & RefCOCO series, Groma-Instruct & 13\% \\ 
M-Understanding & M4-Instruct\cite{li2024llavainterleave} & 16\% \\ 
M-Grounding & \ourtrain\ (Stage-1) & 54\% \\ 
\midrule 
\rowcolor[HTML]{E6E6E6} 
\multicolumn{3}{c}{\textbf{Stage-2}} \\ 
\midrule 
S-Understanding & LLaVA-OV-data & 9\% \\ 
S-Grounding & RefCOCO series, Groma-Instruct & 7\% \\ 
M-Understanding & M4-Instruct\cite{li2024llavainterleave} & 8\% \\ 
\multirow{2}{*}{M-Grounding}  & M-Grounding (Stage-1) & 27\% \\ 
 & M-Grounding (Stage-2) & 49\% \\ 
\bottomrule
\end{tabular}%
}
\caption{Training data proportion for two stages.}
\label{tab:stage1_data_proportion}
\end{table}

\subsection{Synthesizing Free-form MIG Data}
\label{appendix:stage2data}

The algorithm for CLIP adaptive similarity image input is shown in Algorithm \ref{alg:adaptive_image_selection}.
We further display our prompt template for image caption generation, bounding box label refinement and instruction tuning data generation in the following pages and several stage-2 data examples in Figure~\ref{fig:sft}. 

Specifically, we deploy Qwen2-VL-7B for detailed image caption generation and Qwen2-VL-72B for bbox label refinement. The inference process is accelerated through vLLM framework~\cite{kwon2023efficient}.
\begin{algorithm}[t]  
\caption{CLIP Adaptive Similarity Selection}
\label{alg:adaptive_image_selection}
\begin{algorithmic}[1]
\REQUIRE Images $\textbf{I}$, adaptive selection range $k$, $thres \in (0, 1)$
\ENSURE Final Image Set $\textbf{F}$
\STATE Initialize $\textbf{F} \leftarrow \emptyset$  
\STATE Extract $\textbf{F}_I \leftarrow \text{Features of } \textbf{I}$ 
\WHILE{$\textbf{F}_I$ is not empty}
    \STATE Randomly select $thres \sim \text{Uniform}(0.1, 1)$
    \FOR{each $f_i \in \textbf{F}_I$}
        \STATE $s_{ij} = \text{similarity}(f_i, f_j), \forall f_j \in \textbf{F}_I, j \neq i$
    \ENDFOR
    \STATE $\textbf{Sort\_S}_i = \text{Sort}(s_{ij})[1:]$
    \STATE ${k} \leftarrow \lfloor thres \times (\text{len}(\textbf{Sort\_S}_i)) \rfloor$  
    \STATE $\textbf{Candidates} \leftarrow \textbf{Sort\_S}_i[:{k}]$ 
    \STATE Randomly select $r \sim \text{Uniform}(3, 5)$ 
    \STATE $\textbf{Selected} \leftarrow \text{Sample}(\textbf{Candidates}, r)$  
    \STATE Append $f_i$ and $\textbf{Selected}$ to $\textbf{F}$
    \STATE Remove $f_i$ and $\textbf{Selected}$ from $\textbf{F}_I$
\ENDWHILE
\RETURN $\textbf{F}$
\end{algorithmic}
\end{algorithm}

\section{Details of Two-Stage Training}
\label{appendix:twostage}

This section outlines the data proportions and their respective sources for the two training stages, as summarized in Table~\ref{tab:stage1_data_proportion}.

In stage 1, we leverage both single-image and multi-image datasets encompassing general understanding and grounding tasks to comprehensively enhance the model’s capabilities. At this stage, the stage-1 subset from \ourtrain\ constitutes the largest portion of the training data, with a total of 530k examples. The total training examples for stage-1 is 1 million.

In stage 2, the focus shifts to stimulating the model's free-form MIG abilities by integrating all free-form grounding data from \ourtrain. A significant proportion of stage-1 data is also reused to maintain the previously learned abilities. The total number of training examples in this stage is 200k.

\section{Evaluation Implementation}

\paragraph{Polling-based Evaluation}
When directly requiring the model to generate bounding box coordinates for each image, due to their limited multi-image grounding ability and insufficient instruction following ability, the answer obtained in this way is largely unfaithful and mostly unsatisfactory in instruction following, failing to objectively reflecting the real grounding ability of the model. . Empirically, instructing the models to directly generate all bounding box coordinates results in very serious instruction following issue. The models struggle with plausible outputs, with their answers mostly containing only one bounding box or pure text analysis. 

Considering current model's feeble performance, we transform from directly generating all answers to polling every single image, which facilitates definite and objective evaluation.
Empirically, directly generating all the bounding box coordinates for all images results in lower performance. Yet as illustrated in Table \ref{tab:polling}, Migician still demonstrates great robustness to the variation of evaluation format.

\paragraph{V*Bench Evaluation Implementation}
\label{appendix:v_search}
We transform the high-resolution single image grounding task into a MIG challenge. Specifically, we slice a single high-resolution image into multiple sub-images and directly convert the problem into the group grounding task, which first requires the model to perform image-level locating and then ground the target in that specific image. By utilizing the MIG ability of \ours, we can locate the regions relevant to the input question. Afterward, the model combines the identified region with the original image to generate the answer for the input question, achieving high-accuracy results.

\paragraph{70B Scale Models} The performances of three competitive 70B scale models are illustrated in Table \ref{tab:70B} when equipped with single-image CoT. The general effectiveness of CoT framework is tremendous, with the average performance boost at 20 points. Yet even competitive and much larger model like Qwen2-VL-72B (58.70\%) still can't surpass our Migician (60.49\%) in multi-image grounding, demonstrating great competence.

\paragraph{Single-Image Grounding}
As presented in Table 4, Migician not only acquires free-form multi-image grounding capabilities but also demonstrates continual and consistent performance improvements on the RefCOCO series single-image grounding benchmark, surpassing specialized grounding models such as Griffon v2 and GroundingGPT by a large margin. Additionally, Migician outperforms Qwen2-VL-7B in terms of average scores.

\paragraph{Human-Level Evaluation}

\label{appendix:human_eval}

We engaged five human volunteers to answer questions from MIG-Bench. For the static image difference, common object grounding, region locating, and group grounding tasks, we randomly selected 20\% of the test examples for efficient evaluation.

The volunteers are instructed to answer the MIG questions by directly drawing bounding boxes on the images, with the resulting annotations then compared to the ground truth using automatic IoU calculation. To ensure unbiased results, evaluations from different volunteers were conducted separately, preventing mutual distraction or information leakage.

\section{Multi-Task Learning}
Our whole training process involves the learning process of multiple distinct tasks. How does the actual learning efficiency alter compared with learning these tasks separately, can they contribute to each other or comprise to some extent?

We conduct experiments that only expose the model to omni-task dataset and the results are shown in Table \ref{tab:separate}. It clearly reveals the conflicts of learning various tasks, with mixes multi-task training consistently surpassing omni-task learning by a huge margin. When we directly merge the checkpoints of all these trained specialized models~\cite{ilharco2022editing}, the merged model fail at excelling at most tasks, with the average performance falling behind simple multi-task learning.

\section{Case Study}
\begin{figure*}[h] 
    \centering
    \includegraphics[width=0.90\textwidth]{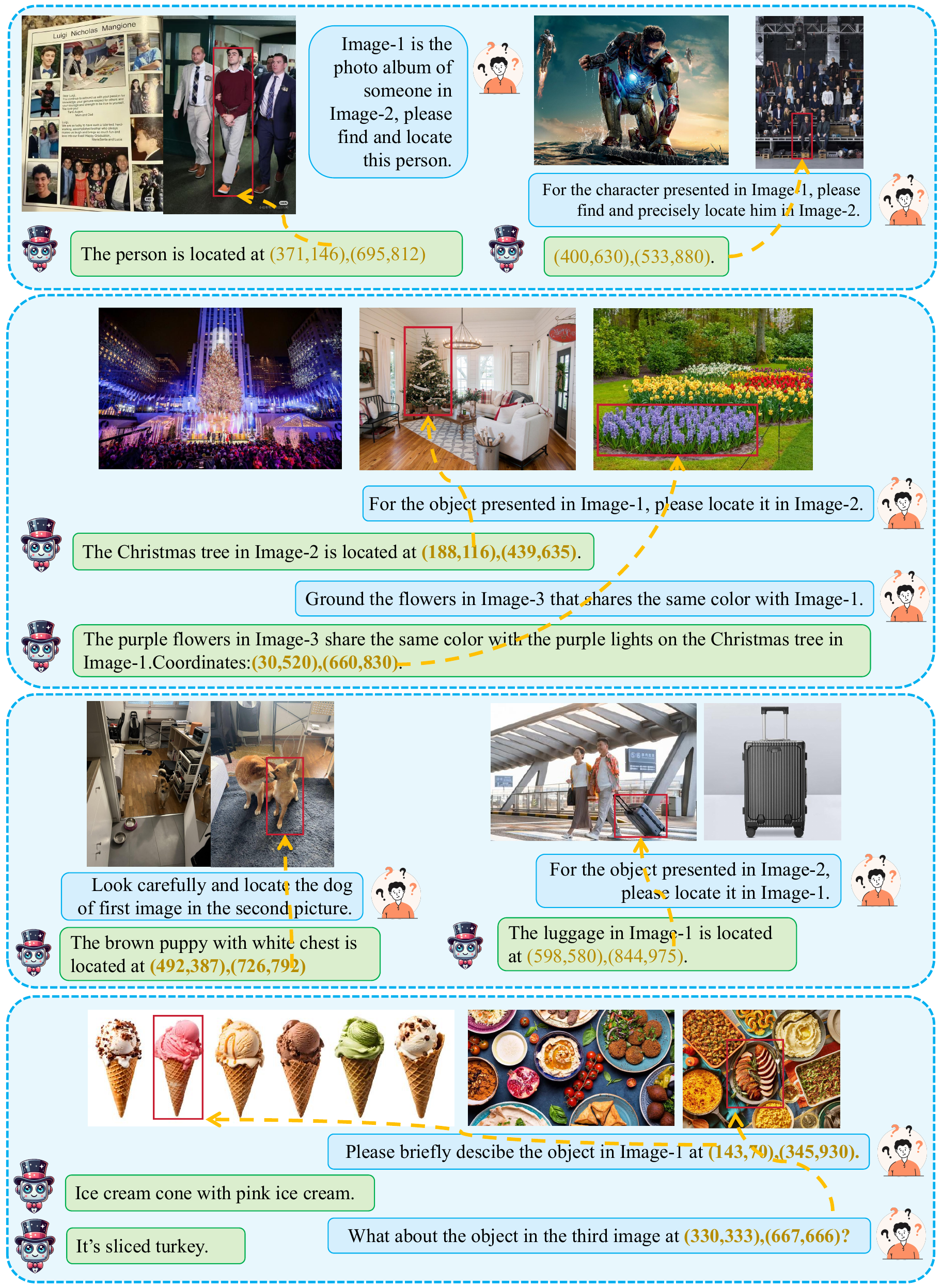} 
    \caption{Example cases of the free-form multi-image grounding ability of Migician.} 
    \label{fig:case1} 
\end{figure*}

\begin{figure*}[h] 
    \centering
    \includegraphics[width=0.98\textwidth]{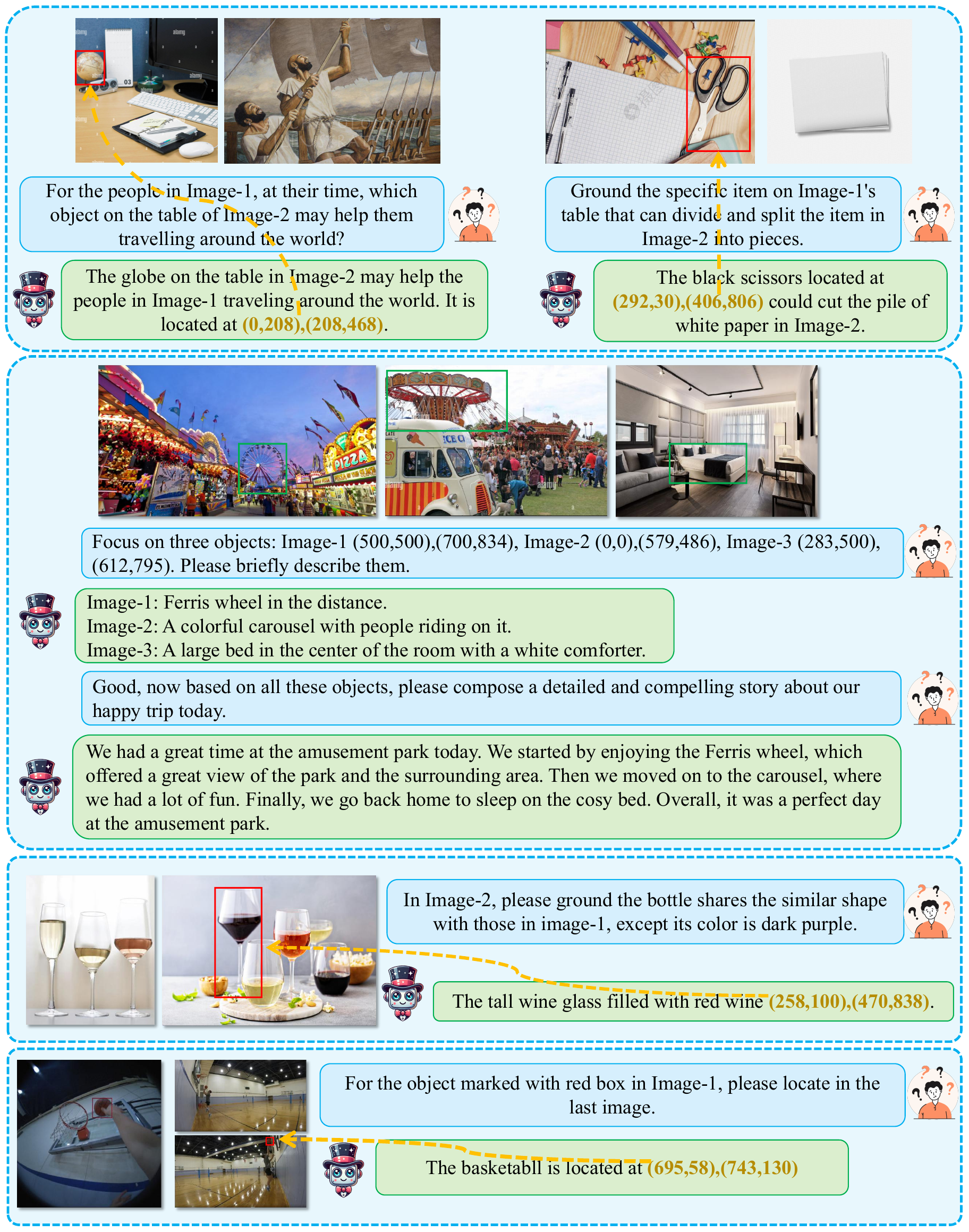} 
    \caption{Example cases of the free-form multi-image grounding ability of Migician.} 
    \label{fig:case2} 
\end{figure*}

\begin{figure*}[h] 
    \centering
    \includegraphics[width=0.95\textwidth]{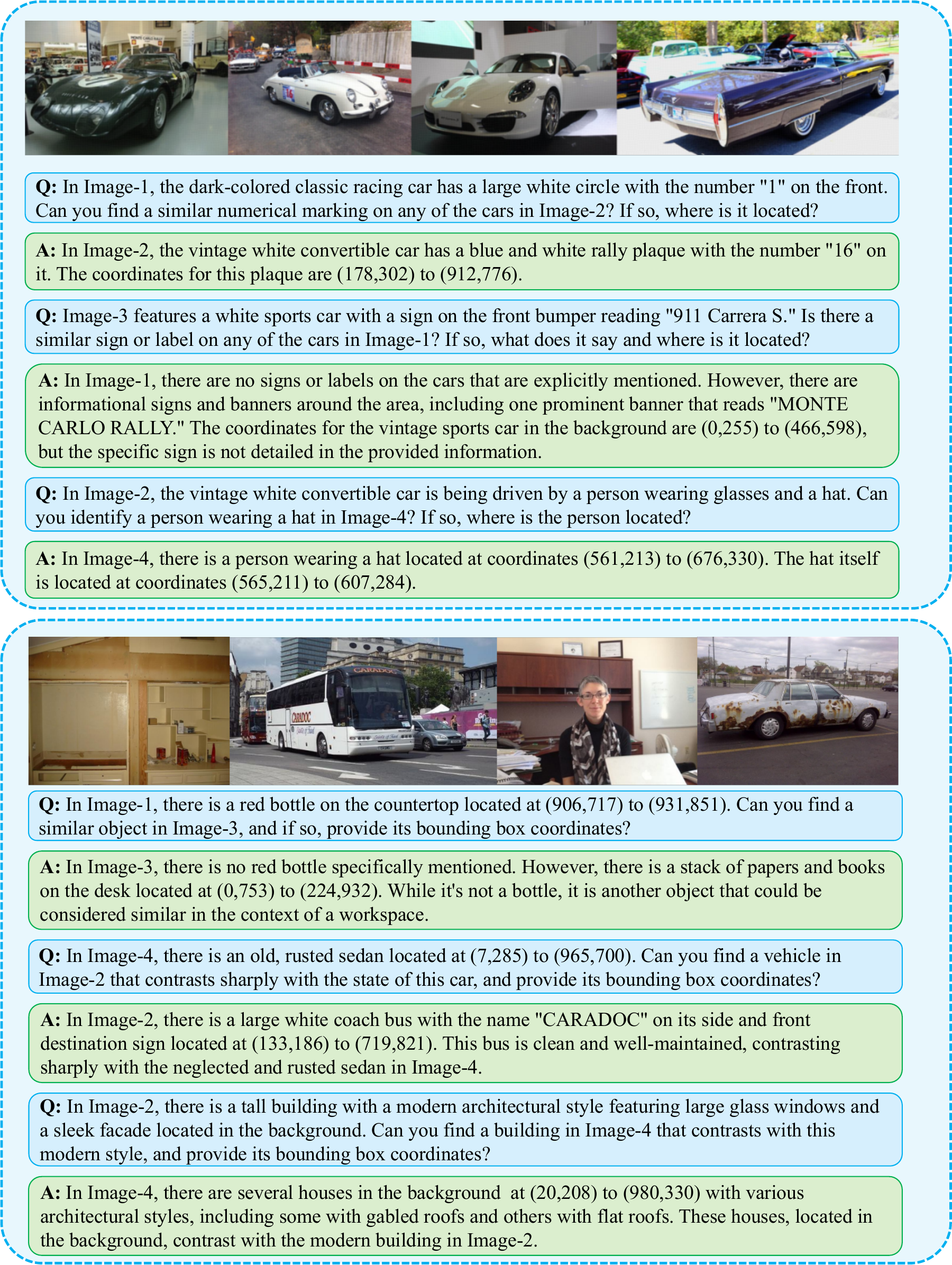} 
    \caption{Training Examples of the free-form instruction tuning data.} 
    \label{fig:sft} 
\end{figure*}

We provide detailed cases comprehensively reflecting the free-form MIG ability of \ours in Figure~\ref{fig:case1}, \ref{fig:case2}, as well as our instruction tuning data details examples in Figure~\ref{fig:sft}.

\begin{tcolorbox}[float*=h!, title=Prompt Template for Single-Image CoT, width=\textwidth, colframe=black]
    \small 
    \textbf{Task: Static diff}
    \\
    Step-1: Compare these two images carefully and tell me where does they differ. Please answer briefly in single phrase or words.\\
    Step-2: According to the object difference/change: [RESPONCE], please ground this difference with bounding box coordinates.
    \\ \\ 
    \textbf{Task: Robust diff}
    \\
    Step-1: Compare these two images carefully and describe the prominent different object with really simple words or phrase.\\
    Step-2: Now ground the object difference/change : "[RESPONCE]"  with bounding box coordinates.
    \\ \\ 
    \textbf{Task: Referring Grounding}
    \\
    Step-1: Watch carefully and briefly describe the object in the Image-1.\\
    Step-2: Please find and ground the object <|object\_ref\_start|>[RESPONCE]<|object\_ref\_end|> with bounding box coordinates.
    \\ \\ 
    \textbf{Task: Common Object}
    \\
    Step-1: These images share one object in common. Recognize it and tell me its name in single phrase or words.\\
    Step-2: Please locate and ground the target object according to the reference: <|object\_ref\_start|> [RESPONCE] <|object\_ref\_end|>
    \\ \\
    \textbf{Task: Region Locating}
    \\
    Step-1: Describe the content of the XXXth picture with simple phrase or words.\\
    Step-2: Please ground the object <|object\_ref\_start|>[RESPONCE]<|object\_ref\_end|> with bounding box coordinates.
    \\ \\
    \textbf{Task: Multi-View}
    \\
    Step-1: Describe the object in the first image marked with red bounding box(<|box\_start|> (A,B),(C,D) <|box\_end|>) with simple phrase or word. You can refer to other images for more precise recognition and description.\\
    Step-2: Locate and ground the object <|object\_ref\_start|> [RESPONCE] <|object\_ref\_end|> with bounding box coordinates.
    \\ \\
    \textbf{Task: Object Tracking}
    \\
    Step-1: Describe the object in the first image marked with red bounding box with simple phrase.\\
    Step-2: Now ground the target moving object [RESPONCE] with bounding box coordinates.
    \\ \\
    \textbf{Task: Group Grounding}
    \\
    Step-1: Just recognize and tell me which image is it in. Answer from: Image1 | Image2 | Image3...\\
    Step-2: [Selected Image] + [Original Question]\\
    \textbf{Note: For group grounding, the single image at step-2 is selected by matching the answer from step-1. If the framework fails to extract the target image, we send the first image by default.}
    \\ \\
    \textbf{Task: Reasoning}
    \\
    Step-1: [Original Question] + Name this object in the Image-2 with simple phrase.\\
    Step-2: Please locate and ground the object <|object\_ref\_start|>[RESPONCE]<|object\_ref\_end|> with bounding box coordinates.
    \\ \\
    \textbf{Task: Correspondence}
    \\
    Step-1: For the first image, describe the semantic/functional feature of the area marked by the red bounding box (<|box\_start|>(A,B),(C,D)<|box\_end|>).\\
    Step-2: Ground the area that shares the same semantic or functional meaning of: [RESPONCE].
    \\ \\
    \textbf{Format Prompt}
    \\
    Format: <|box\_start|>(x1,y1),(x2,y2)<|box\_end|>. Don't generate additional words.\\
    \textbf{Note: we deploy this prompt for better instruction following.}
\end{tcolorbox}

\begin{tcolorbox}[float*=h!, title=Prompt Template for Caption and Instruction Data Generation, width=\textwidth, colframe=black]
    \small 
    \textbf{Bbox Refinement Template}\\
    Now I’d like you to inspect the original image carefully. Then filter, refine and enhance these annotated objects. Finally, just give me your final modified annotations.\\

    *Filtering*\\
    Based on you insightful observation of the image, please eliminate the obviously inaccurate (object,bbox) pairs, which in supposed to be small in quantity.\\
    
    *Refine*\\
    Refine and enhance the original class/name of each object into a short yet richer caption containing its attributes like color, position, feature(e.g plane <|box\_start|>(x1,y1),(x2,y2)<|box\_end|> -> dark gray plane flying in the sky <|box\_start|>(x1,y1),(x2,y2)<|box\_end|>).\\
    
    *Amplify*\\
    If any important objects are missing from the annotations, and you believe they are significant and essential, and you are confident of their location, feel free to add them to the final annotations.\\
    
    *Output Format*\\
    Modified object caption followed by its bounding box coordinates.\\
    
    Now the original bounding box annotations I give to you are:
    \\ \\
    \textbf{Caption Generation}\\
    Describe this image thoroughly in a fluent paragraph. Include all the objects and their attributes(color, shape, size and feature), relative position and relationship.
    \\ \\ 
    \textbf{Multi-image Grounding Instruction Generation}\\ \\
    \textit{Template 1}\\
    Based on the following detailed information of multiple images, please compose meaningful and flexible CROSS-IMAGE grounding questions that link different objects across the images by their attributes similarity/contrast—such as color, position, features, gender, size, shape, etc.—or by other potential logical connection between them.\\
    Specifically:\\
    1.The questions should include CROSS-IMAGE grounding requests that requires the answer to identify and locate various potentially connected object across different images. You can use the connection or similarity between these objects to refer the target item.\\
    2.When referring an object in the question, keep the reference description concise and avoid giving away unnecessary information(like bbox or over-detailed caption) that could lead to answering too easily. You are encouraged to refer the target object to be grounded by the connection of these objects, instead of explicitly point out the object. For instance: “ground the car in image-2 that contrasts most in quality with the shabby vehicle in image-4”, rather than “ground the fancy red sports car(explicitly pointing out) in image-2 that contrasts most in quality with the shabby vehicle in image-4”, by doing so we can also introduce a bit reasoning process.\\
    3.Include the bounding box coordinates of referred object in the answer as well as the explanation. (Actually you can get a lot of information from the coordinates, which are formatted as (x1,y1),(x2,y2))\\
    4.Strictly format the output as simple Q: A:. In answer, follow the format <ref>object</ref> for objects mentioned.
    \\
    Below are the detailed image captions and the objects in the corresponding images:
    \\ \\


\end{tcolorbox}


\end{document}